\documentclass[a4paper, conference]{IEEEtran}
\IEEEoverridecommandlockouts

\usepackage{tabularx}
\usepackage{multirow}
\usepackage{cite}
\usepackage{amsmath,amssymb,amsfonts}
\usepackage{algorithmic}
\usepackage{graphicx}
\usepackage{textcomp}
\usepackage{xcolor}
\usepackage{smartdiagram}
\usepackage{tikz}
\usepackage{hyperref}
\usetikzlibrary{shapes.geometric, arrows ,intersections}

\usepackage{blindtext}
\usepackage[a4paper, bottom=41mm, top=19.1mm, left=13.5mm, right=13.5mm]{geometry}

\tikzstyle{hierarchy} = [rectangle, rounded corners, draw, align=center, top color=white, bottom color=blue!10, minimum width=2.2cm, minimum height=1cm, text width=1.8cm, text centered]
\tikzstyle{arrow} = [thick,->,>=stealth]

\def\BibTeX{{\rm B\kern-.05em{\sc i\kern-.025em b}\kern-.08em
    T\kern-.1667em\lower.7ex\hbox{E}\kern-.125emX}}
\newcommand\copyrighttext{%
  \footnotesize \textcopyright 2022 IEEE. Personal use of this material is permitted.
  Permission from IEEE must be obtained for all other uses, in any current or future 
  media, including reprinting/republishing this material for advertising or promotional 
  purposes, creating new collective works, for resale or redistribution to servers or 
  lists, or reuse of any copyrighted component of this work in other works. 
  DOI: \href{<https://ieeexplore.ieee.org/document/9956031>}{10.1109/ICPR56361.2022.9956031}}
\newcommand\copyrightnotice{%
\begin{tikzpicture}[remember picture,overlay]
\node[anchor=south,yshift=10pt] at (current page.south) {\fbox{\parbox{\dimexpr\textwidth-\fboxsep-\fboxrule\relax}{\copyrighttext}}};
\end{tikzpicture}}%
\begin{document}

\title{3D Face Reconstruction for Forensic Recognition - A Survey}

\author{\IEEEauthorblockN{Simone Maurizio La Cava$^1$, Giulia Orrù$^1$, Tomáš Goldmann$^2$, Martin Drahansky$^2$, Gian Luca Marcialis$^1$}
\IEEEauthorblockA{
\textit{$^1$University of Cagliari (Italy), $^2$Brno University of Technology (Czech Republic)}\\
$^1$\{simonem.lac, giulia.orru, marcialis\}@unica.it, $^2$\{igoldmann, drahan\}@fit.vutbr.cz}}

\maketitle
\copyrightnotice

\begin{abstract}
3D face reconstruction algorithms from images and videos are applied to many fields, from plastic surgery to the entertainment sector, thanks to their advantageous features. However, when looking at forensic applications, 3D face reconstruction must observe strict requirements that still make unclear its possible role in bringing evidence to a lawsuit. Shedding some light on this matter is the goal of the present survey, where we start by clarifying the relation between forensic applications and biometrics. To our knowledge, no previous work adopted this relation to make the point on the state of the art. Therefore, we analyzed the achievements of 3D face reconstruction algorithms from surveillance videos and mugshot images and discussed the current obstacles that separate 3D face reconstruction from an active role in forensic applications. 
\end{abstract}

\IEEEpeerreviewmaketitle

\section{Introduction}\label{sec:intro}

In the last few decades, much attention has been paid to the use of 3D data in facial image processing applications. This technology has shown to be promising for robust facial feature extraction \cite{2D3Dsurvey, robust, robustsurvey}. In uncontrolled environments, it limits the effects of adverse factors such as unfavourable illumination conditions and the non-frontal poses of the face with respect to the camera \cite{robust,uncalibrated}. 

Among the various scenarios, developing personal recognition based on 3D data appears to be a "hot topic" due to the accuracy and efficiency obtainable from  matching faces, thanks to the complementary information of shape and texture  \cite{Afzal2020, 3Dstereo, ICPrecognition}. However, acquiring such data requires expensive hardware; moreover, the enrolment process is much more complex \cite{dis3D,dis3D2,dis3D3,uncalibrated,dis3D4}. Thus, face recognition technology was mainly developed in the 2D domain. The acquisition of 2D images is more straightforward than that of 3D ones, as it does not require specific hardware, but often makes the recognition task challenging due to the significant variability in the facial appearance \cite{uncalibrated}. 3D face reconstruction (3DFR) from 2D images and videos may overcome these limits, combining the ease of acquiring 2D data with the robustness of 3D ones.

One of the possible fields that could benefit from these advantageous characteristics is that of forensics, which often deals with probe images of unidentified people faces in non-frontal view, uncontrolled environments, uncooperative way, such as in the case of the ones captured by CCTV (Closed-Circuit Television) cameras. In such context, it is common to have mugshots, that is, frontal and profile images of subjects routinely captured by law enforcement agencies \cite{Liang2018}. 

From the first attempt of face recognition from mugshots \cite{firstMugRec}, 3D reconstruction techniques were exploited too. However, to be suitable for real-world forensic applications, any system of the kind should satisfy strict constraints leading to the legal validity of the conclusions during a lawsuit or in the investigation phase \cite{JainSurvey, forensicRequirements}. For this reason, it is necessary to analyze the methods which employ 3DFR to shed some light on their admissibility in the forensic scenario. Although other authors investigated the state of the art of 3DFR from 2D images or videos \cite{monocular2018, monocular2021, statistical2020, uncalibrated} and its applications to face recognition \cite{frontal2020, statistical2020, uncalibrated}, none of them considered the requirements they have to satisfy to be potentially employed in such context and how forensics can benefit from their adoption. This paper is a first step to this goal, where we analyze the evolution of this cross-domain field under such perspectives and the novelties introduced to date.

The paper's structure is as follows. Section \ref{sec:biometrics} analyzes the relation between forensics and biometrics, facial traits in particular. The state of the art assessment of 3DFR methods for face recognition from mugshot images is reported in Section \ref{sec:mugshot}. A review of other proposed forensic-related applications of 3DFR from facial images and videos is carried out in Section \ref{sec:other}.
Finally, Section \ref{sec:conclusion} discusses how all the aspects above converge in a unified view.

\section{Face recognition and forensics}\label{sec:biometrics}

\begin{figure}[!t]
\centerline{
\begin{tikzpicture}[
    thick,
    >=stealth',
    dot/.style = {
      draw,
      fill = blue,
      circle,
      inner sep = 0pt,
      minimum size = 4pt
    }
  ]
  \coordinate (O) at (0,0);
  \draw[->] (0,0) -- (8,0) coordinate (xmax);
  \draw[->] (0,0) -- (0,4.25) coordinate (ymax);
  \path[name path=x] (0.3,0.5) -- (6,4);
  \path[name path=y] plot[smooth] coordinates {(-0.3,2) (2,1.5) (4,2.8) (6,4)};
  \scope[name intersections = {of = x and y, name = i}]
    \draw[blue!30, ultra thick] plot coordinates {(0,0) (0.5,0.25) (2,0.25) (2.5,1) (4,1) (4.5,1.75) (5.5,2.25) (6,2.25) (6.5,2.75) (7,3)};
    \draw[blue!30, ultra thick] plot coordinates {(0.5,3.75) (1,3.75)};
	\node[black] at (2.05,3.75) {\tiny Total number of publications};

	\node[black] at (7.8, -0.25) {\scriptsize Year};
	\node[black] at (-0.25, 4) {\scriptsize N};

    \draw[black!] plot coordinates {(0.5,0.25) (0.5,0.75)}; 
	\node[black] at (0.5,1.05) {\tiny Gallery };
	\node[black] at (0.5,0.9) {\tiny enlargement};

    \draw[black!] plot coordinates {(2.5,0.5)(2.5,1.5)}; 
	\node[black] at (2.5,1.8) {\tiny Video-based };
	\node[black] at (2.5,1.65) {\tiny reconstruction};
	\node[black] at (2.5,0.35) {\tiny Probe/set };
	\node[black] at (2.5,0.2) {\tiny normalization};

    \draw[black!] plot coordinates {(4.5,1.75) (4.5,2.25)}; 
	\node[black] at (4.5,2.5) {\tiny Multi-reference};
	\node[black] at (4.5,2.35) {\tiny reconstruction};

    \draw[black!] plot coordinates {(5.5,1.75) (5.5,2.25)}; 
	\node[black] at (5.5,1.6) {\tiny DL-based};
	\node[black] at (5.5,1.45) {\tiny recognition};

    \draw[black!] plot coordinates {(6.5,2.75) (6.5,3.25)}; 
	\node[black] at (6.5,3.5) {\tiny DL-based shape};
	\node[black] at (6.5,3.35) {\tiny reconstruction};

   \draw[black!] plot coordinates {(7,3) (7,2.5)}; 
	\node[black] at (7,2.35) {\tiny Generation of};
	\node[black] at (7,2.2) {\tiny CCTV-like images};

    \foreach \Point in {(0,-0.05), (0.5,0.20), (2.5,0.95), (4.5,1.7), (5.5,2.2), (6.5,2.7), (7,2.95)}{
        \node[blue!80] at \Point {{\textbullet}};
    }
    \foreach \Point\year in {(0,-0.2)/2007, (0.5,-0.2)/2008, (1,-0.2)/2009, (1.5,-0.2)/2010, (2,-0.2)/2011, (2.5,-0.2)/2012, (3,-0.2)/2013, (3.5,-0.2)/2014, (4,-0.2)/2015, (4.5,-0.2)/2016, (5,-0.2)/2017, (5.5,-0.2)/2018, (6,-0.2)/2019, (6.5,-0.2)/2020, (7,-0.2)/2021}{
        \node[black!80] at \Point {{\tiny\year}};
    }
    \foreach \Point\n in {(-0.2,0.5)/2, (-0.2,1)/4, (-0.2,1.5)/6, (-0.2,2)/8, (-0.2,2.5)/10, (-0.2,3)/12, (-0.2,3.5)/14}{
        \node[black!80] at \Point {{\tiny\n}};
    }
  \endscope
\end{tikzpicture}
}
\vspace{-0.3cm}
\caption[justification=centering]{The cumulative number of publications (N) proposing methods for the employment of 3DFR to enhance forensic recognition.}
\label{fig:mugevo}
\vspace{-0.55cm}
\end{figure}
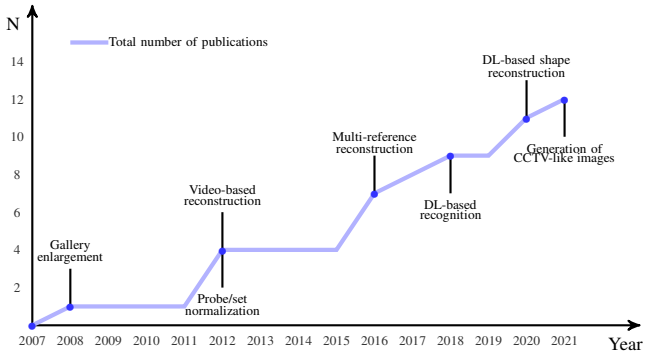

The face represents one of the most valuable clues in many criminal investigations due to its advantageous characteristics with respect to other biometrics \cite{recognitionsurvey, faceInvestigation} and the growing number of surveillance cameras \cite{surveillanceMarket}.
Although both biometric recognition and forensic identification seek to link evidence to a particular individual \cite{JainRoss2015}, research in these fields has been pursued independently for many years due to their different goals. Figure \ref{fig:mugevo} shows the amount of papers per year, and points out the increasing interest of scholars, but also the difficulties in achieving significant scientific contributions (numbers are noticeably larger in other 3DFR applications \cite{uncalibrated}). 

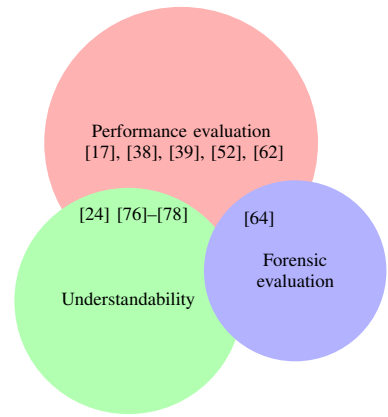
\begin{figure}[!t]
\centering
\begin{tikzpicture}
  \begin{scope}[blend group = soft light]
    \fill[red!30!white]   (90:1.7) circle (1.8);
    \fill[green!30!white] (210:0.8) circle (1.5);
    \fill[blue!30!white]  (0:1.5) circle (1.2);
  \end{scope}
  \node [font=\scriptsize, text width=3cm, align=center] at (90:1.7) {Performance evaluation \\\cite{Dutta2012, vanDam2012, Rahman2016, Liang2018, Liang2020}};
  \node [font=\scriptsize] at (210:0.8)   {Understandability};
  \node [font=\scriptsize, text width=1.5cm, align=center] at (0:1.5)   {Forensic evaluation};
  \node [font=\scriptsize, align=center] at (132:1)    {\cite{HanJain2012} \cite{Zhang2008, Zeng2016, Zeng2017}};
  \node [font=\scriptsize] at (35:1.2)    {\cite{vanDam2016}};
\end{tikzpicture}
\vspace{-0.2cm}
\caption[justification=centering]{Taxonomy of forensic recognition methods based on 3DFR with respect to the evaluation levels for forensic purposes (based on \cite{forensicRequirements}).}
    \label{fig:workseval}
\vspace{-0.55cm}
\end{figure}

Techniques and systems designed for biometrics, especially the automated ones, are appealing for their potential in addressing some forensic domain's problems in a more efficient and standardized way \cite{YanOsadciw2008, JainRoss2015, faceforensics, forensicExaminers, localmarks}.
However, to bridge the gap between the two domains, it should be necessary to establish a robust methodology for forensic recognition based on statistical and probabilistic methods providing guidelines for quantifying biometric evidence value and its strength based on assumptions, operating conditions, and uncertainty implicit in the casework \cite{Tistarelli2014}. Therefore, a set of interpretation methods should be defined independently of the baseline biometric system and integrated into the considered algorithm \cite{Neumann2012, Tistarelli2014}, allowing to reach conclusions in court trials in agreement with three constraints (figure \ref{fig:workseval}): performance evaluation, understandability, and forensic evaluation \cite{JainSurvey, linkage, forensicRequirements}. Performance evaluation concerns the basic trust level of the system and its performance for a specific purpose; therefore, it supports the forensic practitioner's decision when using such a system to perform a given task. For instance, a biometric system could be considered suitable for a specific task whenever it is tested and achieves good performance on ground truth data which are representative of data on which such system is employed (e.g., a face recognition system which performs well on good quality frontal images could not achieve the same performance on images acquired by CCTV cameras with different head poses) \cite{forensicRequirements}. 
Understandability (or interpretability) is the ability of a human to understand the functioning of a system, its purpose, its features, as well as its output and the inferences made which lead to that result. In particular, its evaluation supports the decision of whether the outcome of the system is suitable. In order to be understandable, a biometric system must be explainable, meaning that it has to make its functioning and its purpose clear to legal decision-makers (e.g., judges) who are, typically, not experts in these topics \cite{JainSurvey, assessment, forensicRequirements}. 
Forensic evaluation is the assignment of a relative plausibility of information over a set of competing hypotheses (or "propositions") and supports the forensic practitioner's opinion regarding the level of confidence and the weight of evidence, output of the automated system, when reaching a decision \cite{automatedFaces, assessment, forensicRequirements}. The system's performance and understandability are taken into account in forensic evaluation, together with contextual information and general knowledge either included in the decision process or formalized into the automated system itself \cite{forensicRequirements}. From a technical perspective, a biometric system must ease the interpretation of results in the forensic context through assessments as likelihood ratio values \cite{ENFSI, faceslikelihood, Tistarelli2014} rather than binary results as in traditional biometric applications \cite{automatedFaces}. 

Coherently with these aspects, the conditions for applying biometrics and, specifically, the face in forensic recognition rely on the characteristics of acquired data. Firstly, they should meet minimal requirements in terms of "quality" \cite{YanOsadciw2008}. Although not defined in a rigorous way, the "quality" refers to factors that lead to blurriness, distortion, and artifacts in images. They may be caused by (1) the camera employed, whose sensor, optic, and analog-to-digital converter impact on the image resolution, the dynamic of gray-levels, its ability to focus-on-the-target, (2) environmental conditions such as the illumination and the background of the scene, the same weather conditions (rainy/cloudy), (3) the subject's distance from the camera that adds scaling and out-of-focus problems, his/her camouflage to evade recognition (sunglasses, beard/moustache, hat/cap, makeup), (4) the image processing embedded into the camera or next to the raw data acquisition, such as compression and re-sizing  \cite{lowquality,YanOsadciw2008,recognitionsurvey,faceslikelihood,ENFSI}. Secondly, the data amount is crucial from the viewpoint of the classification system to be trained and fine-tuned \cite{YanOsadciw2008}, yielding to the creation of large-scale databases for the evaluation of face recognition algorithms (e.g., \cite{LFW}).

During the investigation phase, the subject's identity is unknown, and the possible identities within a suspect reference set need to be rendered and sorted \cite{YanOsadciw2008} in terms of likelihood with respect to the evidence (e.g. a frame captured from a CCTV camera) \cite{Ali2014}. 
In addition to the classic challenges related to facial recognition in uncontrolled environments (such as low resolution, large poses, and occlusions \cite{GuoZhang2019}), forensic recognition faces other challenges, such as acquisition systems which are set up cheaply and subjects that actively try not to be captured by cameras, enhancing the previously cited issues and introducing novel problems such as heavy compression, distortions, and aberrations due to imperfection in cameras constructions \cite{Zeinstra2018}.
Thanks to its greater representational power than 2D facial data, 3DFR can alleviate some of these problems since 3D data provides a representation of the facial geometry which is invariant to pose and illumination. 
Depending on the characteristics of the probe image and of the reference set narrowed down by police and forensic investigation, whenever the investigator is required to compare these images and it is necessary or advantageous to use an automatic face recognition system, 3DFR can be employed by following two different approaches, namely a view-based and model-based approaches, to improve the performance of facial recognition systems and, therefore, enhancing its admissibility in legal trials.

\begin{figure}[!t]
\centering
\begin{tikzpicture}
      [sibling distance=13em, level distance=6.1em,
      every node/.style={shape=rectangle, rounded corners, draw, align=center, top color=white, bottom color=blue!10, minimum width=2.7cm},
      level 2/.style={sibling distance=8.5em, level distance=6.7em}]
      \node[minimum height=1cm]{\scriptsize \textbf{Enhancement} \\ \scriptsize \textbf{approach}}
        child{node[minimum height=2.2cm]{\scriptsize \textbf{Model-based approach} \\ \scriptsize Reconstruction from \\ \scriptsize probe-like data \\ \scriptsize (e.g., \cite{ForenFace}) for \\ \scriptsize identification tasks}
            child{node[minimum height=1.3cm]{\scriptsize \textbf{Probe} \\ \scriptsize \textbf{frontalization} \\ \scriptsize \cite{HanJain2012}}}}
        child{node[minimum height=2.2cm]{\scriptsize \textbf{View-based approach} \\ \scriptsize Reconstruction from \\ \scriptsize mugshot-like data (e.g.,\\ \scriptsize \cite{ColorFERET, PIE, MultiPIE, PCSO}) \\ \scriptsize for verification tasks}
            child{node[minimum height=1.3cm]{\scriptsize \textbf{Gallery} \\ \scriptsize \textbf{enlargement} \\ \scriptsize \cite{Zhang2008, HanJain2012, Liang2018, Liang2020}}}
            child{node[minimum height=1.3cm]{\scriptsize \textbf{Gallery} \\ \scriptsize \textbf{adaptation} \\ 
            \scriptsize \cite{Dutta2012, Zeng2016, Zeng2017}}}};
\end{tikzpicture}
\vspace{-0.1cm}
\caption[justification=centering]{Taxonomy of performance enhancement methods through 3DFR for forensic recognition.}
    \label{fig:apptax}
\vspace{-0.55cm}
\end{figure}
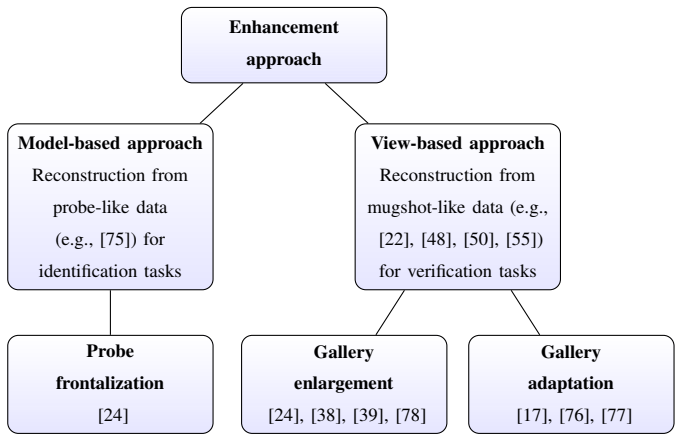

In a view-based approach, the reference set is adapted to the probe image, such that it matches the pose of the represented face \cite{view}. Although it allows comparing facial images under similar poses, this approach requires a reference set containing images of suspects captured in such pose or synthesizing such view through the 3D model of each suspect. In the latter case, each 3D model can be adapted after applying a pose estimation algorithm on the probe image before employing the actual recognition system \cite{Dutta2012, Zeng2016, Zeng2017}. Another proposed strategy is to introduce a gallery enlargement phase instead, which consists of projecting the 3D model in various predefined poses in the 2D domain to enhance the representation capability of each subject and then employing the synthesized images in the recognition task \cite{Zhang2008, HanJain2012, Liang2018, Liang2020}. However, the view-based approach represents a suitable choice whenever multi-view face images of suspects are captured during enrollment for the purpose of highly accurate authentication, such as in the case of the verification task in face recognition \cite{HanJain2012}, although it usually involves higher computational cost both in terms of time and memory with respect to the model-based counterpart.

In a model-based approach, the adaptation phase is performed on the probe image to synthesize a face in frontal view through the 3D face model reconstructed from the probe image itself \cite{HanJain2012}. The normalized (or "frontalized") face is then matched to the frontal faces within the gallery set to determine the subject's identity in the probe image, thus reducing the variability which the recognition system must address \cite{Hassner2015, Ferrari2016}. This approach is suitable for real-world scenarios in which it is necessary to seek the identity of an unknown person within a probe image or video in a large scale mugshot database \cite{HanJain2012}, as in a so-called face identification task in biometric recognition, for maximizing the likelihood of returning the potential candidates while minimizing the processing for the following verification task.
Despite the generally lower computational cost, this approach is only applicable when it is possible to synthesize good quality frontal view images with the original texture since it could provide complementary information for recognition with respect to the shape \cite{ICPrecognition, Liu2018}, thus presenting minimum quality requirements for the probe images, which is not often the case of real forensic scenarios. Furthermore, it could be necessary to handle the consequent texture artefacts in the resulting frontal image \cite{Hassner2015, Cao2020, RotateAndRender}.

Therefore, although it could be possible and convenient to employ approaches and methods designed for biometrics in some forensic cases involving face recognition, these should meet specific requirements to be admissible in legal trials. 

\section{3D face reconstruction for mugshot-based recognition}\label{sec:mugshot}

Although many attempts have been performed in the past years for reconstructing faces in the 3D domain either from a single image or multiple images of the same subject, only a few were evaluated for their potential applications in forensics. One of the most promising and explored approaches employs 3DFR for enhancing facial recognition from forensic mugshot images captured by law enforcement agencies. Therefore, in this section, we review the proposed methods based on such an approach, analyzing their advantages and drawbacks, and whether they satisfy the previously seen criteria for their potential admissibility in forensic cases.

To our knowledge, the earlier study on 3DFR from mugshot images for forensic recognition was proposed in 2008 by Zhang et al. \cite{Zhang2008}, who employed a view-based gallery enlargement approach to recognize probe face images in arbitrary view with the aid of a 3D face model for each subject reconstructed from mugshot images.
To reconstruct the shape of such a model, they proposed a multilevel variation minimization approach that requires a set of facial features, named landmarks \cite{landmarks, landmarkclasses}, specified on a pair of frontal-side views to be used as constraining points (i.e., eyes, eyebrows, nose profiles, lips, ears, and points interpolated between them \cite{controlpoints}). Finally, they recovered the corresponding facial texture through a photometric method. 
They evaluated their approach in based face recognition experiments on the CMU PIE database \cite{PIE}, through a holistic face matcher \cite{eigenfaces} and a local one \cite{LBP}, restricting the rotation angles of the probe images to ±70°. This analysis revealed a significant improvement in average recognition accuracy with respect to the original mugshot gallery, especially when the rotation angle of the face in the probe image is larger than 30°.
However, the limit of the rotation angle of faces in probe images and the use of traditional face matchers rather than state of the art ones do not allow to assess the actual improvement in the effectiveness of 3DFR from mugshot images in terms of forensic recognition \cite{HanJain2012, Liang2018}. Other drawbacks of the proposed method are the possible artefacts caused by the assumed model \cite{Zeng2016} and the poorly explored image texture. Furthermore, although results suggested enhanced performance, they performed the analysis on a small-scale database containing only 68 subjects. Finally, despite the improvement in performance and the use of a local face matcher that enhances understandability, the authors did not employ any method for easing the forensic evaluation.

Four years later, Han and Jain \cite{HanJain2012} proposed a 3DFR method from a pair of frontal-profile views based on 3DMM (3D Morphable Model) \cite{3DMM}, a generative model for realistic face shape and appearance, to aid the reconstruction process. 
They reconstructed the 3D face shape and the correspondence of landmarks between frontal and profile images, while they extracted the texture by mapping the facial image to the 3D shape.
They employed their method under a face recognition perspective through a gallery enlargement approach and a probe frontalization approach. They evaluated them on subsets of PCSO \cite{PCSO} and FERET databases \cite{ColorFERET} through a local face matcher and a commercial one, revealing an improved recognition accuracy in both cases.
One of the most evident limits of the reconstruction approach in a forensic context is that the involved 3DMM is a global statistical model which is limited in recovering facial details \cite{uncalibrated}, as it could be dominated by the mean 3D face model, which potentially introduces a bias of the outcome towards the underlying model \cite{vanDam2016}. This aspect could be further enforced by the relatively low quality of the images contained in the FERET database. Furthermore, the involved 3DMM could cause evident distortion when the model is largely rotated \cite{Zeng2017, Liang2020}. Other limits of this work are that the authors did not fully explore the texture and that they did not use state of the art face matchers \cite{Zeng2016, Liang2018}. Therefore, as in the previous case, despite the improvement in performance while using a local face matcher that enhances understandability, the authors did not employ any framework for easing the forensic evaluation of their method.

In the same year, Dutta et al. \cite{Dutta2012} proposed a method based on 3DFR for improving face recognition from non-frontal view images through a view-based gallery adaptation approach.
They applied existing recognition systems on the 16 common subjects in the CMU PIE \cite{PIE} and Multi-PIE \cite{MultiPIE} databases, containing frontal images and surveillance images, respectively. 
This method could be particularly advantageous whenever a poor quality surveillance system has acquired the analyzed probe image while, at the same time, it is possible to acquire mugshot images of the suspects in the same pose or by reconstructing the 3D face from images having higher quality. However, this approach requires that it is possible to accurately estimate the pose of the face in the probe image.
Furthermore, the small number of subjects involved in the study should be enlarged to simulate a forensic case and evaluate the improvement entity for assessing their applicability in real-case scenarios. Despite the advantages in some application contexts in terms of performance, the authors did not take into account understandability or forensic evaluation.

Similarly, Zeng et al. \cite{Zeng2016, Zeng2017} reconstructed 3D faces from frontal, left, and right 2D forensic mugshot images through multiple reference models to obtain more accurate outcomes for enhancing recognition performance through a gallery adaptation approach. To this aim, they used a coarse-to-fine 3D shape reconstruction approach based on the three views through a photometric method and multiple reference 3D face models. The use of multiple reference models is an attempt to limit the homogeneity of reconstructed 3D face shape models and increase the probability of finding the most similar candidate for the single parts of the input face. 
The so-reconstructed 3D face shapes were then used in the recognition task to establish correspondence between the local semantic patches around seven landmarks on the arbitrary view probe image and those on the gallery of mugshot face images, assuming that patches will deform according to head pose angles. 
The authors \cite{Zeng2016} tested their approach on the CMU PIE \cite{PIE} and Color FERET \cite{ColorFERET} databases, showing how their method can improve performance with respect to the non-deformed semantic patches \cite{LBP}, to a commercial face recognition system, and even to the method proposed by Zhang et al. \cite{Zhang2008}. The authors \cite{Zeng2017} also evaluated the enhancement using a machine learning (ML) classifier on different poses within the Bosphorus \cite{Bos} and Color FERET \cite{ColorFERET} databases.
As the authors suggested, the improvement in recognition capability from arbitrary position face images is due to the more robustness of semantic patches to pose variation and the higher inter-class variation introduced by the subject-specific 3D face model.
A limitation of this work is the out-of-date involved face matchers \cite{Liang2018}. Furthermore, although the method employs multiple reference models, the outcome could still be unwillingly biased toward them \cite{vanDam2016}. Finally, despite the fact that the proposed method enhances the performance of an understandable recognition approach, authors did not perform any forensic evaluation.

In 2018, Liang et al. \cite{Liang2018} proposed an approach for arbitrary face recognition based on 3DFR from mugshot images which fully explores image texture.
The proposed shape reconstruction approach is based on cascaded linear regression from 2D facial landmarks estimated in frontal and profile images. After reconstructing the 3D shape, they approached the texture recovery through a coarse-to-fine approach. 
Therefore, they employed the proposed method in a recognition task on a subset of images from each subject of the Multi-PIE database \cite{MultiPIE}, through a gallery enlargement approach on state of the art matchers based on deep learning (DL). Furthermore, they compared the performance first and after the gallery enlargement and by fine-tuning the matchers with the generated multi-view images. The results highlighted improved recognition accuracy in large-pose images, especially with fine-tuned matchers. In particular, this method provides better results than the one proposed by Han and Jain \cite{HanJain2012}, probably because of the major focus on the reconstruction of texture information \cite{Liang2018}.
Hence, the most significant novelties introduced by this work are the textured full 3D faces reconstructed from the mugshot images and the analysis on DL-based matchers, inherently more robust to pose variations than traditional ones \cite{Liang2018}. Furthermore, they fine-tuned those matchers with the enlarged gallery, revealing even better performance than through the previous gallery enlargement approaches.
A limit of the proposed method is that it does not consistently work across all pose directions, revealing worse performance for some poses than in the original gallery (e.g., in frontal pose). Furthermore, the authors did not take into account understandability or forensic evaluation.

In 2020, the same authors published an extension of this work \cite{Liang2020}, in which they also proposed a DL-based shape reconstruction.
In this work, the authors extended the evaluation of the face recognition capability of the proposed method based on linear shape reconstruction by employing a subset of the Color FERET database \cite{ColorFERET}, obtaining a higher recognition accuracy on average as in the case of the Multi-PIE database \cite{MultiPIE}.
Furthermore, they tried to solve the drawback of their previous work, related to worse recognition performance for some poses, with respect to usage of the original gallery, through a fusion between the similarity scores obtained by both the original mugshot images and the synthesized ones. This approach, evaluated on the Multi-PIE database \cite{MultiPIE}, revealed consistently better performance on all the pose angles.
Despite the proposed novelties, the authors did not assess if the proposed DL-based shape reconstruction approach is able to enhance recognition capability. Finally, the study did not consider understandability or forensic evaluation.

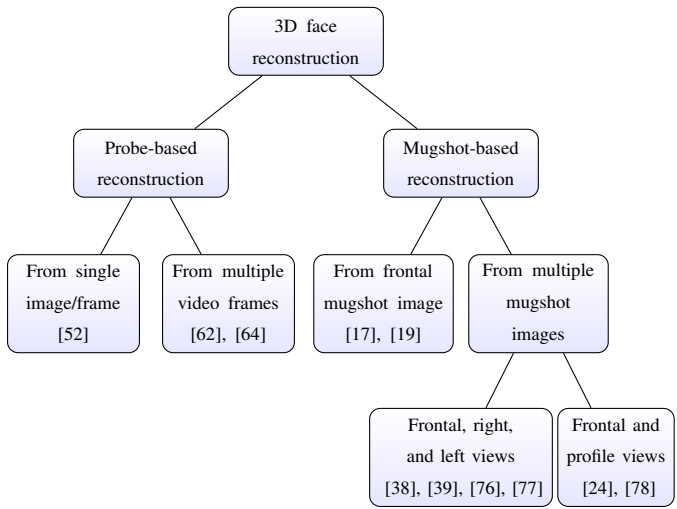
\begin{figure}[!t]
\centering
\begin{tikzpicture}
      [sibling distance=11.6em, level distance=4.6em,
      every node/.style={shape=rectangle, rounded corners, draw, align=center, top color=white, bottom color=blue!10},
      level 2/.style={sibling distance=5.8em, level distance=5.3em},
      level 3/.style={sibling distance=5.8em, level distance=5.8em}]
      \node[minimum height=0.9cm, text width=1.8cm]{\scriptsize 3D face reconstruction}
        child{node[minimum height=0.9cm, text width=1.8cm]{\scriptsize Probe-based reconstruction}
            child{node[minimum height=1.3cm, text width=1.5cm]{\scriptsize From single image/frame \cite{Rahman2016}}}
            child{node[minimum height=1.3cm, text width=1.5cm]{\scriptsize From multiple video frames \cite{vanDam2012, vanDam2016}}}}
        child{node[minimum height=0.9cm, text width=1.8cm]{\scriptsize Mugshot-based reconstruction}
            child{node[minimum height=1.3cm, text width=1.6cm]{\scriptsize From frontal mugshot image \cite{Dutta2012, Ferkova2020}}}
            child{node[minimum height=1.3cm, text width=1.6cm]{\scriptsize From multiple mugshot images}
                child{node[minimum height=1.3cm]{\scriptsize Frontal, right, \\ \scriptsize and left views \\ \scriptsize \cite{Zeng2016, Zeng2017, Liang2018, Liang2020}}}
                child{node[minimum height=1.3cm]{\scriptsize Frontal and\\ \scriptsize profile views \\ \scriptsize \cite{Zhang2008, HanJain2012}}}
            }};
\end{tikzpicture}
\vspace{-0.2cm}
\caption[justification=centering]{Taxonomy of 3DFR approaches in forensic scenarios.}
    \label{fig:rectax}
\vspace{-0.55cm}
\end{figure}

A quantitative comparison among the previously reviewed methods would require the usage of the same face matchers and their evaluation on the same ground truth data, and this is often unfeasible due to many factors, such as the current state of the art databases when the work has been proposed. However, a qualitative comparison is provided in the latter section (Section \ref{sec:conclusion}), allowing us to obtain a general overview of the state of the art 3DFR for forensic recognition.

\section{Other applications of 3D face reconstruction in forensics}\label{sec:other}

In addition to recognition from mugshot images, 3DFR could represent a valuable aid to forensics in other contexts as well in which it is possible to facilitate the recognition of a subject. An example of forensic application related to such context which would benefit from the properties of a 3D face model is the search for missing persons. Taking into account such a scenario, Ferková et al. \cite{Ferkova2020} proposed a method that includes demographic information to improve the outcome of the reconstruction from a single frontal image and, at the same time, speed up the related computation. In particular, starting from an image of the missing person's face, the method estimates the related 3D shape taking into account age, gender and similarity between the landmarks of the reference depth images and the ones previously annotated in the input image. Then, planar meshes are generated by triangulating between the input image and the depth image. 
Despite the good geometrical results, the width of the outcome is usually overstretched and the generated 3D face model does not include the forehead.
Furthermore, the authors did not quantitatively evaluate the contribution of their method to recognition capability or their potential admissibility in forensic scenarios.

Similarly to some of the previous studies, Rahman et al. \cite{Rahman2016} highlighted how 3D face models could enhance forensic recognition from CCTV camera footage. In particular, they reconstructed the 3D face models from single frames by optimizing an Active Appearance Model (AAM), an algorithm that matches a statistical model of object shape and appearance to an image \cite{ABC}. Therefore, they evaluated the improvement in the recognition capability of different ML models with respect to 2D AAMs. However, this study on the possible application of 3DFR to forensic recognition from surveillance videos only represents a preliminary analysis due to the use of a database related to a limited number of subjects and that is not provided as well. Finally, the authors did not assess if their method improves the performance of state of the art models and its potential admissibility in forensic recognition in terms of understandability and forensic evaluation.

With a similar purpose, van Dam et al. \cite{vanDam2012} proposed a method based on a projective reconstruction of landmarks on the face and an auto-calibration step to obtain the 3D face model from CCTV camera footage, taking into account the specific case of the fraud to an ATM with an uncalibrated camera.
The method proposed by the authors reconstructs the facial shape utilizing landmarks estimated from multiple frames and an auto-calibration method \cite{autocalibration}.
The authors also analyzed how the quality of the resulting 3D face model is affected by the number of frames and noise on the landmarks, assessing the minimum values for an acceptable 3D face shape.
However, they did not quantitatively assess if the proposed integration of 3DFR could improve recognition performance with respect to its 2D counterpart. Finally, the authors did not assess the proposed method's understandability and forensic evaluation.

In 2016, the same authors proposed another method to reconstruct a 3D face from multiple frame images for an application in the forensic context \cite{vanDam2016}. Such a method employs a photometric method to estimate both texture and 3D shape of the face without employing any model to avoid generating an outcome biased towards any face model, therefore inherently enhancing the suitability in a forensic face comparison process.
The proposed method is a coarse-to-fine shape estimation process that firstly provides a coarse 3D shape \cite{vanDam2013} and other pose parameters from landmarks in multiple frames, and then a refined shape by assessing the photometric parameters for every point in the 3D model. The last step also allows estimating texture information, then providing the dense 3D face model.
The authors evaluated the proposed method in a recognition task on the database they acquired, composed of single-camera video recordings of 48 people, each containing frames with different facial views. To this aim, they compared the reconstructed textures with the ground truth images through FaceVACS \cite{FaceVACS}, by increasing the considered frames among iterations, revealing enhancement in recognition results in most cases. Furthermore, using the likelihood ratio framework, the authors highlighted that in more than 60\% of the cases, data initially unsuitable for forensic cases became meaningful in the same context through the proposed method.
As the authors suggested, the outcomes can be used to generate faces under different poses, while are not suitable for shape-based 3D face recognition.
Despite the enhanced suitability in forensic scenarios, one of the most significant drawbacks of the proposed approach is that the model-free reconstruction approach is computationally more burdensome than a model-based one and requires multiple images. Furthermore, the authors did not quantitatively evaluate their method on publicly available databases. Although the authors did not assess understandability, they introduced a forensic evaluation of their method based on 3DFR.

Unlike all previous approaches, Loohuis \cite{Loohuis2021} proposed to employ 3DFR for facing the lack of facial images, which could be used in training ML and DL models for face recognition tasks in forensic settings, such as in a surveillance scenario. Therefore, the author combined a method for the generation of face images with rendering techniques to simulate such adverse conditions and assessed the impact of the resulting synthetic images in existing face recognition systems. In particular, the employed 3DFR method \cite{Deng2019}, based on a DL model and a 3DMM \cite{Basel}, was applied on the single images of a subset of the ForenFace database \cite{ForenFace} to generate images simulating different levels of image degradation.
This represents a preliminary study since the method does not perform well on very low quality images. However, a reasonable level of degradation that is present in many forensic scenarios can still be mimicked, as shown by the comparable recognition performance on the generated images and the reference ones.

\section{Discussions}\label{sec:conclusion}

In this paper, we reviewed the state of the art of 3D face reconstruction (3DFR) from 2D images and videos for forensic recognition, evaluating the proposed approaches with respect to the requirements of a potential forensics-related system. Furthermore, the proposed approaches for enhancing forensic recognition in terms of performance were analyzed together with their potential application scenarios (figure \ref{fig:apptax}).

The previously described studies mainly focus on enhancing performance of recognition tasks in different contexts, such as the identification (or verification) of suspects within a gallery of mugshot images \cite{Zhang2008, HanJain2012, Dutta2012, Zeng2016, Zeng2017, Liang2018, Liang2020} or the search for missing persons \cite{Ferkova2020}.
Despite the promising results, most of the previously described studies did not evaluate their methods considering other requirements of an automated system supporting forensic analysis, related to understandability and forensic evaluation \cite{JainSurvey, forensicRequirements}. However, some of them implicitly used a face recognition algorithm based on local descriptors \cite{Zhang2008, HanJain2012, Zeng2016, Zeng2017}, which supports the understandability of the output \cite{localmarks, interpretable}. Furthermore, a single study \cite{vanDam2016} employed a framework for easing forensic evaluation. Although many of them improved recognition performance by enhancing, in particular, the robustness to pose variations in various probe and reference data types, underlining the related application scenarios (figure \ref{fig:rectax}), none of them considered all the aspects that would support judicial suitability. Moreover, the proposed methods do not assess their robustness to some typical issues of forensic cases, such as the presence of occlusions \cite{realocclusion, challenges}.

Although most of the proposed methods aim to enhance face recognition performance, they are not comparable quantitatively due to the variability in the considered settings. One of the most relevant differences among them is related to the involved databases, which differ in terms of acquisition environment, numerosity, availability, and quality of data. Those differences are partly caused by the state of the art of such databases when the studies were proposed. In particular, future studies should be based on databases suitable for forensic research, such as the ForenFace database \cite{ForenFace}, as they take realistic circumstances into account. Furthermore, they should evaluate the face reconstruction accuracy on large-scale 3D face databases, such as the FIDENTIS database \cite{FIDENTIS}. 

Another factor that should be considered while designing a system based on 3DFR is the eventual 3D reference face model, as outcomes of the system could be biased toward such a model \cite{JainSurvey}, making it unsuitable in forensic cases \cite{vanDam2016}. Therefore, despite the computationally demanding process, a model-free reconstruction approach should be employed, such as the stereophotogrammetry, which allows capturing craniofacial morphology in high quality \cite{3Dstereo}. However, a drawback of these approaches is the requirement of multiple images of the suspects \cite{vanDam2016}, which cannot be acquired in any forensic case.

Future studies in the analyzed field should focus on enhancing the recognition capability in forensic scenarios through 3DFR, which may help improve performance in extremely unfavourable conditions, typically encountered in criminal cases, by considering both shape and texture in state of the art recognition systems.
In particular, they should be evaluated on data representative of forensic trace and reference material, also considering the robustness to other common factors altering the appearance such as facial hair and the presence of occlusions.
In our opinion, one of the priorities of the research community should concern more about the understandability of the algorithms underlying the proposed approaches, both in terms of reconstruction and recognition. They may play a central role in the effective integration of 3D face reconstruction from 2D images and videos in the forensic field. Similarly, the employment of frameworks for easing forensic evaluation by non-expert professionals should become a practice for stressing the admissibility of the proposed methods in real cases.
Although the research in this field is still incomplete and there are still many issues to be studied, it revealed promising results for its future involvement in real-world applications, and we hope that this survey can be a first step to provide helpful guidelines towards the realization of such scenario.

\section*{Acknowledgment}
This work is supported by the Italian Ministry of Education, University and Research (MIUR) within the PRIN2017 - BullyBuster - A framework for bullying and cyberbullying action detection by computer vision and artificial intelligence methods and algorithms (CUP: F74I19000370001).

\end{document}